# ND-MLS: A novel method for data augmentation and its effect

**Wen Yang [1], Rui Wang[1], Yanchao Zhang [1], \***

[1]    School of Mechanical Engineering and Automation, Zhejiang Sci-Tech University, Hangzhou, 340018, China;

\*    Correspondence: yczhang@zstu.edu.cn

**Abstract:** Data augmentation greatly increases the amount of data obtained based on labeled data to save on expenses and labor for data collection and labeling. We present a new approach for data augmentation called nine-dot MLS (ND-MLS). This approach is proposed based on the idea of image deformation. Images are deformed based on control points, which are calculated by ND-MLS. The method can generate over 2000 images for one existing dataset in a short time. To verify this data augmentation method, extensive tests were performed covering 3 main tasks of computer vision, namely, classification, detection and segmentation. The results show that 1) in classification, 10 images per category were used for training, and VGGNet can obtain 92% top-1 acc on the MNIST dataset of handwritten digits by ND-MLS. In the Omniglot dataset, the few-shot accuracy usually decreases with the increase in character categories. However, the ND-MLS method has stable performance and obtains 96.5 top-1 acc in ResNet on 100 different handwritten character classification tasks; 2) in segmentation, under the premise of only ten original images, DeepLab obtains 93.5%, 85%, and 73.3% m_IOU(10) on the bottle, horse, and grass test datasets, respectively, while the cat test dataset obtains 86.7% m_IOU(10) with the SegNet model; 3) with only 10 original images from each category in object detection, YOLO v4 obtains 100% and 97.2% bottle and horse detection, respectively, while the cat dataset obtains 93.6% with YOLO v3. In summary, ND-MLS can perform well on classification, object detection, and semantic segmentation tasks by using only a few data.

## 1. Introduction

Given a large number of labeled samples, deep neural networks are very good at solving classification problems in computer vision. However, it is difficult to obtain a large number of labeled data sets in many scenarios. Therefore, many scholars have struggled to solve this problem under conditions with insufficient samples[1]. Although fine-tuning a pretrained network is prone to overfitting[2], it is impossible to obtain satisfactory results with sufficient samples. With fine-tuning, the necessary sample size is not as small as that of humans, who need only a few samples to learn to categorize. Many samples are inevitably needed to train the network due to the millions of parameters being updated. A large amount of sample training is also required to acquire prior knowledge even using one-shot image categorization methods[3]. For example, Siamese neural networks only need a small number of samples, and the recognition accuracy far exceeds the human level on similar objects. Siamese neural networks have achieved great success in face recognition problems[4]. However, such networks also need a large amount of data to learn prior knowledge[5]. Second, they can only achieve good results in the classification of similar objects[6]. From this perspective, fine-tuning or short-learning are premised on massive data training. Thus, we propose effective data augmentation to further enlarge the size of the training dataset, which can satisfy the data needs of the network training.

Data augmentation is usually used to reduce overfitting in the process of training[7]. It is the most common method to rapidly expand datasets. Data augmentation approaches such as random cropping[8], flipping[7], and random erasing[9] have long been used in model training. Currently, most studies are devoted to architectural innovation. It is essential to consider the data management method to achieve the desired results using less data. Thus, the data acquisition and preprocessing expenses can be reduced. For example, although the cost of acquiring images is low, target labeling requires extensive manual labor, which leads to a sharp increase in costs. In practice, the situation is complex and changeable, and it is difficult for a model to cover all situations. It is a good idea to quickly augment the dataset through appropriate methods to significantly lower the requirement of the training dataset and speed up the advancement of machine learning projects.

Let us review the commonly used data augmentation methods, which can be divided into three categories: ①Traditional data augmentation methods, such as flipping, scaling and rotating, which were used at the very beginning of deep learning. Flipping[7] avoids the influence of the object position in the image. Scaling[7] can eliminate the influence of the size of the recognition object. Changing the pixel value of the RGB channel[2] on the image should be performed carefully. In many cases, the color of an object cannot be casually changed. For example, identification of a traffic light is very sensitive to color. Increasing noise[7] is also a good way to make the trained model adapt to more complex environments. Last but not least, changing the background[10] can increase the versatility of the dataset since the targets are placed in various environments. ② The GAN-based method is an emerging method of data augmentation, although it has rarely been used. Before generating new images, a large number of images are required for training, and there is no clear standard to evaluate the quality of the generated image[11]. As a result, it is not appropriate to directly use a large amount of data generated by GAN for training because the generated data are likely to negatively affect training. Third, 3D-based data augmentation[12] is another newly introduced method that enlarges the dataset from different angles of the targets' 3D model. It requires a ready-made 3D model, but very few studies have one; thus, it is difficult to simulate the complexity of reality through 3D models.

In this paper, a nine-dot moving least squares (ND-MLS) method was proposed for data augmentation, and the effectiveness of this ND-MLS data augmentation method in classification, positioning, and segmentation was verified. In this research, only ten manually labeled samples are needed to train the model to reach the goal of object detection and segmentation. A set of handles should be selected to control the ND-MLS-based deformation. Multiple points need to be reasonably selected to control the deformation of the object, and it will be discussed in detail how to perform the selection in different situations. By controlling point selection, the dataset can be rapidly expanded to meet the demands for fine-tuning the amount of data.

To verify the effectiveness of the proposed method, first, the expanded dataset was used in classification tasks. After data augmentation by the ND-MLS method, various classification network frameworks were used to test the performance of data set augmentation. Second, the expanded dataset was used in the task of object detection and object segmentation. After using the ND-MLS method to rapidly expand the dataset, LeNet, AlexNet, VGGNet, ResNe, and MobileNet were selected to verify the performance of ND-MLS on the classification task, SegNet, PspNet, and DeepLab on the semantic segmentation task, and Faster RCNN, YOLOv3, and YOLOv4 on the semantic segmentation task.

Data augmentation is an important process in deep neural model training. To realize the change of an image, users need to select a set of handles, which may take the form of lines[13], polygon grids [14], or points[15,16] provides an image deformation method to manipulate images of real-world objects, which can minimize the amount of local scaling and shear. Image deformations using moving least squares do not produce globally smooth deformations such as triangulates. Image deformation means that the shape of objects can be manipulated by some control points[15]. By moving the limited control points on the image, the rest of the image can be adjusted automatically under certain rules. The change in the control points' positions only affects a small area of the image near the point. For example[16], by moving some control points, image deformation produces complex expression changes such as joy, anger, mourning, and happiness. Changes in character movements can also be implemented. This method is mostly used in the production of movies and animation to make the transitions in videos more realistic. It will be notable to introduce the idea of image deformation to data augmentation.

A research hotspot in recent years has been the use of a few existing training samples to train models. ND-MLS can achieve the same effect as few-shot learning[17]. Few-shot learning is mostly carried out from the perspective of neural network structure models. Generally, the few-shot learning model is first trained with a large number of samples in the early stage, and only a few samples are needed to learn new things. Humans are very good at few-shot learning, and scholars have developed a variety of models to imitate the ability of humans. [18] proposed methods to learn new things quickly by using augmented memory capacities in the model. The Siamese network for one-shot image recognition[3] is based on metric learning to train the model. Few-shot learning is basically based on the innovation and improvement of the model, as well as taking the human nature of learning new things into consideration. Compared to humans, CNNs barely perform as well as humans under data insufficiency.

In the verification section, computer vision tasks, including classification, detection and segmentation tasks, are usually used to verify the effect of ND-MLS augmentation.

Classification has been a major issue in machine learning. Many studies on deep neural network-based classification have been published. LeNet is the earliest CNN mode for handwritten digit recognition[19], and its structure is relatively concise. AlexNet adopts a deeper network, introduces data augmentation, and introduces dropout to avoid overfitting[2]. The emergence of AlexNet has promoted the prosperity of convolutional neural networks[20]. VGG-Nets was proposed by VGG (Visual Geometry Group) of Oxford University, using smaller convolution kernels to make the network deeper[21]. ResNet introduces a residual structure to solve the problem of gradient disappearance, making the network layers more than 100[22]. MobileNet[23] introduces depthwise separable convolution to greatly reduce the size of the model. These networks are generally representative of classification problems, and ND-MLS augmentation will be verified in these frameworks one by one.

Meanwhile, deep neural network-based semantic segmentation has drawn a large amount of attention. SegNet[24] obtained a segmentation network by modifying the VGG-16 network[21]. PspNet[25] aggregates the context information through the proposed pyramid scene parsing network based on different regions to mine global context information and alleviate the lack of information in the global scene in the FCN model[26]. DeepLabV3 + introduced the encoder-decoder to fuse multiscale information[27]. The spatial pyramid pooling module (SPP)[28] and encoding and decoding structure[29] are also used to mine multiscale context content information. At the same time, DeepLabV3+ balances accuracy and time consumption by cavity convolution.[27] SegNet, PspNet and DeepLabV3 + are classical segmentation modes, which will verify the effectiveness of ND-MLS.

Moreover, computer vision-based object detection has long been studied and has been widely used in industry. Deep neural networks have promoted it to another level. This approach can be divided into two types: two-stage detection and one-stage detection[20]. Faster RCNN[30] is a representative two-stage method that improves RCNN[31], and the detection accuracy and speed are satisfactory. YOLOv3[32] and YOLOv4[33], as one-step methods, have great advantages in terms of detection speed. RCNN, YOLOv3 and YOLOv4 were used to validate ND-MLS performance on objection detection.

## 2. Materials and Methods

This paper focuses on the discussion of the Rigid Deformation of Moving Least Squares, in which sets of control points should be selected to control the direction of deformation. Let $p$ be a set of control points and $q$ be the deformed positions of the control points $p$, so it is key to determine the number and position of $p$ and the position of $q$ after it has moved. First, the moving least square method is usually used in image deformation[16]. As shown in formula (1), $f_r$ can produce smooth image deformations by using moving least squares when deformed handles $q$ are the same as $p$, and $f_r$ is an identity function. $V$ is the pixel area of the image, and the new pixel area can be obtained after $f_r$ transformation.

$p*$ and $q*$ are weighted centroids in formula (1). As shown in formula (2), these weights ($w_i$) need to ensure that the value at $p*$ is only affected by the nodes in the subdomain near $p$. This subdomain is called the influence region of $p$, and the nodes outside the influence region have no influence on the $p$ value. Each node in the affected region has a different influence on the value at $p$.

Therefore, the weight within the influence region can be defined as formula (3). The influence of control **points p and q** on image deformation can be changed by changing the value of α. However, when α ≤ 1, the function $f_r$ will lose the property of being smooth everywhere. Therefore, this α value is generally greater than 1.

$$f_r(v) = |v - p^*| \frac{\overrightarrow{f_r}(v)}{\left|\overrightarrow{f_r}(v)\right|} + q^* , \tag{1}$$

$$p^* = \frac{\Sigma_i w_i p_i}{\Sigma_i w_i}, q^* = \frac{\Sigma_i w_i q_i}{\Sigma_i w_i} , \tag{2}$$

$$w_i = \frac{1}{|p_i - v|^{2\alpha}} , \tag{3}$$

In formula (1), the vector $f_r$ is a rotated and scaled version of the vector $\overrightarrow{f_r}(v)$, and $|v - p*|$. As shown in formulas (4,5,6), $A_i$ depends only on $\hat{q}_i$, which can be precomputed before moving control point **q**. Although the method is slow due to normalization, the deformation is quite realistic and almost feels as if the user is manipulating a real object, which greatly ensures the quality of the new samples being generated

$$\overrightarrow{f_r}(v) = \sum_i \hat{q}_i A_i , \tag{4}$$

$$A_i = w_i \begin{pmatrix} \hat{p}_i \\ -\hat{p}_i^{\perp} \end{pmatrix} \begin{pmatrix} v - p^* \\ -(v - p^*)^{\perp} \end{pmatrix}^T , \tag{5}$$

$$\hat{p}_i = p_i - p^*, \quad \hat{q}_i = q_i - q^* , \tag{6}$$

### 2.1. ND-MLS in classification

It is most important to find the positions of the control points **p** and **q** by deforming the object using the moving least square method. First, the nine-point MLS method is adopted for the classification problem, taking handwritten number classification as an example for illustration. As shown in Figure. 1, the coordinates of the first point in the upper right corner are (a, b). The points are arranged at equal intervals, so it is convenient to calculate the position coordinates of each point. However, the initial coordinates of the first point need to be set manually to adjust the coordinate parameters to further simplify and introduce the proportional coefficient $k_p$ in formulas (7,8). Therefore, nine-point coordinates can be easily obtained. As shown in Table 1, *l, w, and h* are the photo length, width and height, respectively, which are the attributes of the photo itself and do not need to be set manually. The position of point p can be determined only by adjusting the $k_p$ value. The $k_p$ value is adjusted according to the size of the actual detection object, and the adjusted point should be as close as possible to the edge of the contour.

$$a = k_p * w , \tag{7}$$

$$b = k_p * h , \tag{8}$$

**Table 1.** The calculation formula of **P**

| $(k_p*w, k_p*h)$ | $(w/2, k_p*h)$ | $(w*(1-k_p), k_p*h)$ |
|---|---|---|
| $(k_p*w, h/2)$ | $(w/2, h/2)$ | $(w*(1-k_p), h/2)$ |
| $(k_p*w, h*(1-k_p))$ | $(w/2, h*(1-k_p))$ | $(w*(1-k_p), h*(1-k_p))$ |

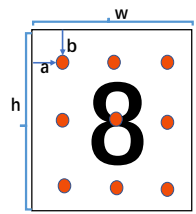 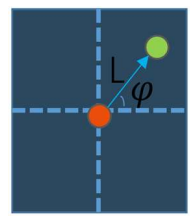

(a)                  (b)

**Figure 1.** the method of handle on MLS. (**a**) the schematic diagram of ND-MLS $p$ point selection; (**b**) the moving direction diagram of $p$ point.

After the control point $p$ is determined, the position of the $q$ point needs to be selected. As shown in Figure 1.b, the position of point $q$ is determined by angle $\varphi$ and length $L$. Formulas (9, 10, 11) are as follows: the initial angle $\varphi$ and the step length $L$ of each angular movement are used to determine where point $q$ can move. For example, in this experiment, $\varphi$=45, $k_s$=90, and $i$=[0,360*$k_s$] ($i$ being a positive integer is improper.). The larger ($k_s$) is, the more types of deformation there are. and then the question about the value of $L$ arises because $p$ and $q$ points should avoid overlapping, and it is necessary to limit the range of $L$. As shown in formulas (12,13), to make it easier to get the value of $L$ within the range, a new parameter $k_l$ was defined to keep $k_l$ between 0 and 1.

Therefore, the nine-point method only needs to adjust four parameters $k_p$, $k_l$, $k_s$ and $\varphi$ to realize various deformations. $\Phi \in (0,360)$,$k_p$, $k_l$, $k_s \in (0,1)$. As shown in formulas (14) and (15), the $k_l$ value determines the range of deformation. The larger the $k_l$ value is, the larger the deformed area. If the image is sensitive to deformation, this value should be smaller. $k_p$ determines where the image needs to be deformed, and $k_s$ determines the number of generated deformed images. The greater the $k_s$ value is, the more images will be generated by deformations.

The number of generated images is related to the number of $p$ points and $k_s$ values. When the number of selected points is determined to be 9 points, only $k_s$ values need to be adjusted. Note that $p$ can be any other number, but 9 is considered to be the most concise one. If $p$ is larger, the redundancy is big. If $p$ is smaller, fewer deformed images are generated, and the augmentation ability becomes poor.

$$Q(x) = p_x + L * cos\varphi \,, \tag{9}$$

$$Q(y) = p_y + L * sin\varphi \,, \tag{10}$$

$$\varphi_i = \varphi_0 + 360 * i * k_s \,, \tag{11}$$

$$L < min\left(k_P * (w - 0.5), k_P * (h - 0.5)\right) \,, \tag{12}$$

$$L = k_l * min\left(k_P * (w - 0.5), k_P * (h - 0.5)\right) \,, \tag{13}$$

$$Q_{x_i} = P_x + L * cos(\varphi_0 + 360 * i * k_s) \,, \tag{14}$$

$$Q_{y_i} = P_y + L * sin(\varphi_0 + 360 * i * k_s) \,, \tag{15}$$

### 2.2. ND-MLS in object detection

In the object detection task, as shown in Figure. 2a, the bounding box can determine its position and its size using ($x_0$, $y_0$, $w_0$, $h_0$) coordinates. ($x_0$, $y_0$) are the coordinates of the lower right corner of the bounding box, and $w_0$ and $h_0$ are the box's height and width, respectively. When the image is deformed, the position of the bounding box will also change, as shown in Figure. 2bc. The position of the new box can be obtained by calculation using a formula (17-20). However, without considering the effect of deformation, it is also difficult to accurately calculate the bounding box after using MSL. According to the deformed box position, it is calculated to estimate the position of the object. As shown in Figure. 2c, it is likely to be different from the position between the deformed box and object. Therefore, the method of directly taking points on the bounding box is not recommended. Taking points on the edge of the ND-MLS object can solve these shortcomings very well, as shown in Figure. 3b, and this approach will be described in detail below.

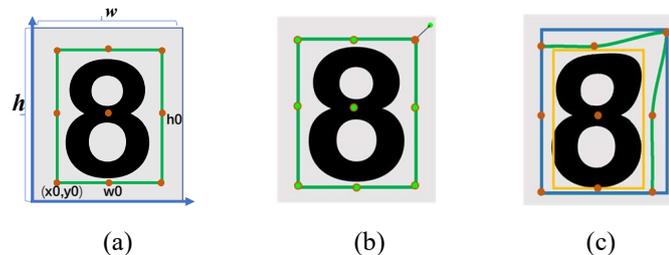

(a)          (b)          (c)

**Figure 2.** (a) original images; (b) schematic diagram of the handles of ND-MLS; (c) ND-MLS deformation images.

$$W_0 = max(Q_y) - min(Q_y) \,, \tag{17}$$

$$H_0 = max(Q_x) - min(Q_x) \,, \tag{18}$$

$$X = min(Q_x) \,, \tag{19}$$

$$Y = min(Q_y) \, , \tag{20}$$

**Note:** $\quad Q_y = \{Q_{y1}, Q_{y2}, Q_{y3} \cdots Q_{yi}\} \, ,$

$$Q_x = \{Q_{x1}, Q_{x2}, Q_{x3} \cdots Q_{xi}\} \, ,$$

## 2.3. ND-MLS in Semantic segmentation

Semantic segmentation is the classification of each pixel in an image, and the object of the image needs to be marked out according to the contour, so it is convenient to obtain the contour information from the label.

According to the label file, it is convenient to obtain the mask of the image. As shown in Figure. 3a, the object marked in the mask can be regarded as a connected area. According to formula (21,22), we can find the center of the barycenter of this area, $f(u, v)$ is the gray value of the pixel with coordinates $(u, v)$, $\Omega$ is the set of connected component regions, and $u$, $v$ is the center of the barycentric coordinates.

Taking the center of the barycenter as the center to establish a rectangular coordinate, we can determine a linear equation that goes through the center of the barycenter in the formula (23). The outer contour of the connected region of the image is composed of many pixel points, which are recorded as set $C$ in the formula (24). It is certain that there must be some intersection between the outer border and the line passing through the barycenter, and these barycentric points are recorded as point set $P$. Considering that the contour of the image is actually composed of discrete points rather than continuous curves, it is impossible to solve the intersection point by using formula (23) directly. The actual intersection point of line $L$ and the contour of the object may not be in point set $C$. It is impossible for us to take some points from point set $C$ since they are close to the actual intersection point. As shown in formula (25), when taking different angles $\beta$, different intersection points can be obtained, and these barycentric points will form point set $p$. The coordinates of point $q$ are determined by the step length $L$ and rotation angle $\varphi$, as shown in eqs. 26 & 27. After image deformation, the mask of the image will also change. To obtain the mask deformation, $v$ in eq. 1 will be replaced with the pixel value of the mask image.

$$\overline{u} = \sum_{(u,v)\in\Omega} u f(u,v) \Big/ \sum_{(u,v)\in\Omega} f(u,v) \, , \tag{21}$$

$$\overline{v} = \sum_{(u,v)\in\Omega} v f(u,v) \Big/ \sum_{(u,v)\in\Omega} f(u,v) \, , \tag{22}$$

$$Y - \overline{v} = tan\beta * \left(X - \overline{u}\right) \, , \tag{23}$$

$$C \cap L = P, (X, Y) \in L \, , \tag{24}$$

$$\left|(Y-v)/(X-u) - tan\beta\right| k \le \xi(|Y_{i+1} - Y_i| > 1 \,\&\, |X_{i+1} - X_i| > 1) \, , \tag{25}$$

$$q(x) = \boldsymbol{p}_x + L * cos(\varphi) \, , \tag{26}$$

$$q(y) = \boldsymbol{p}_y + L * sin(\varphi) \, , \tag{27}$$

The parameters in the moving least square transformation do not need to be calculated again, so the deformed mask can be obtained quickly. After obtaining the new mask, the deformed area can form a bounding box, whose corner coordinates can be used to represent the object location in the object detection task, as shown in Figure. 3b. By selecting the $p$ and $q$ points, image deformation can be carried out to rapidly expand the dataset.

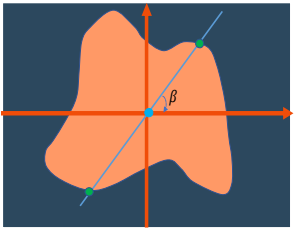

(a)

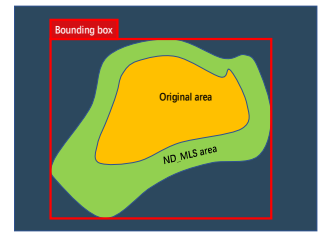

(b)

**Figure.3**. (a) the schematic diagram of ND-MLS *p* point selection in segmentation task; (b) the schematic diagram about bounding box of ND-MLS area in object detection task.

## 3. Results

This section evaluates ND_MLS in three tasks, namely, classification, object detection, and semantic segmentation, to demonstrate the effectiveness of ND-MLS augmentation.

### 3.1. ND-MLS performance in classification tasks

By using $k_p$ of 0.23, $k_l$ of 0.14, and adopting $\varphi$=45°, and further setting $k_s$ as 1/4, 2004 new images are generated by the ND-MLS method for each image. Figure. 4 shows how nine-dot MLS (ND-MLS) is applied to the test image. The deformed images seem to be human written characters. For classification, multiple experiments were performed to explore the relationship between the original sample size and the quality of the ND-MLS data set. Five models, LeNet, AlexNet, VGGNet, ResNet, MobileNet, and 2 open data sets were used to verify the effectiveness of the ND-MLS method.

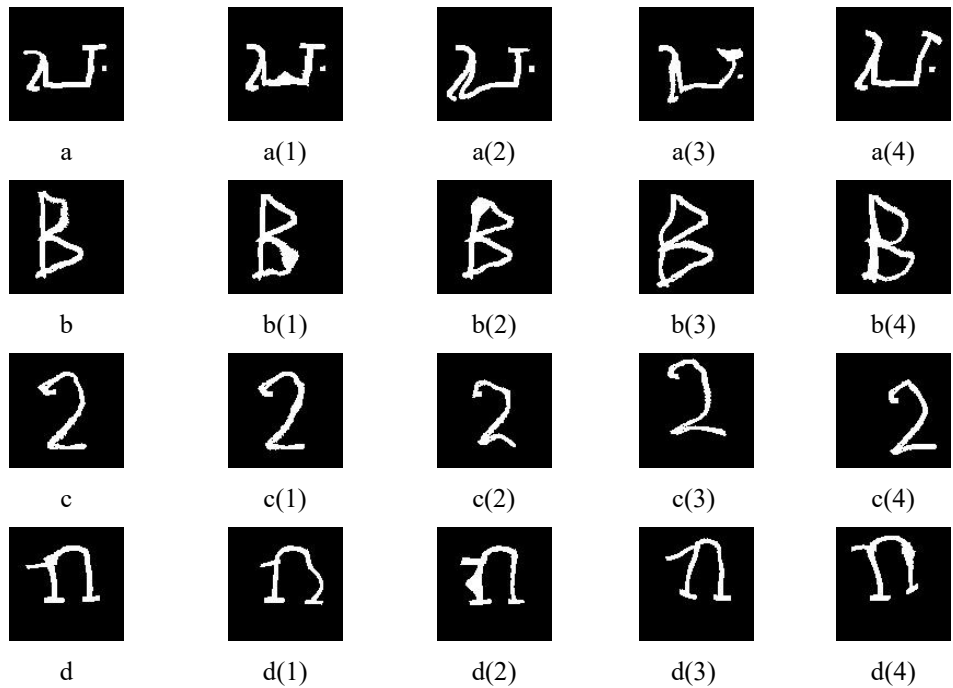

a  a(1)  a(2)  a(3)  a(4)

b  b(1)  b(2)  b(3)  b(4)

c  c(1)  c(2)  c(3)  c(4)

d  d(1)  d(2)  d(3)  d(4)

**Figure 4.** Original images (a, b, c, d) and its deformation using the ND-MLS method (a (1~4), b (1~4), c (1~4), d (1~4)). After deformation, the handwritten characters seem realistic.

#### 3.1.1. MNIST and Analysis

The MNIST dataset of handwritten digits is a commonly used dataset. It consists of 10 classes, namely, 0~9. One hundred images were taken from each class for augmentation, and the rest were used as the test set. Each image can generate 2004 new images by the ND-MLS method, so 2 million images were used for the training model. The test result on the 60,000 test images was reported by the server. The top-1 error rates were used to evaluate this effect.

This is the first study to evaluate the influence of the number of initial images on model accuracy. The results in Table 2 show that the number of original images is positively correlated with the model's accuracy. Let n be the number of original images for each class. When $n$=1, the top-1 test errors are all relatively high. However, the ND-MLS augmentation of $n$=5 performed the $n$=1 best with 1.6% to 40%. Figure 5 reveals the trends of test accuracy when using different numbers of ND-MLS initial images. It is worth noting that the accuracy of the model increases rapidly when n is less than 20. Meanwhile, even when the training samples are the same, the models may perform quite differently. The accuracy of VGGNet is good but fluctuates greatly when n is less than 20, while other models rise relatively steadily. In addition to VGGNet, LeNet has the best performance in 5-shot handwritten number classification.

Finally, we compare $n$ = (10,20,100), leading to 0.08, 0.067 and 0.029 top-1 test errors in Table 2. The results show that ND_MLS augmentation can be used for handwritten numeral classification tasks.

**Table 2.** The top-1 error is on the test set of mnist dataset and reported by the test server. $n$ is the number of original images each class.

|  | $n$=1 | $n$=5 | $n$=10 | $n$=15 | $n$=20 | $n$=100 |
|---|---|---|---|---|---|---|
| LeNet | 0.415 | 0.116 | 0.111 | 0.084 | 0.079 | 0.044 |
| AlexNet | 0.423 | 0.229 | 0.128 | 0.076 | **0.067** | 0.034 |
| VGGNet | 0.556 | **0.016** | **0.08** | **0.01** | 0.068 | **0.029** |
| ResNet | **0.4** | 0.173 | 0.128 | 0.092 | 0.1 | 0.049 |
| MobileNet | 0.5 | 0.239 | 0.163 | 0.118 | 0.106 | 0.039 |

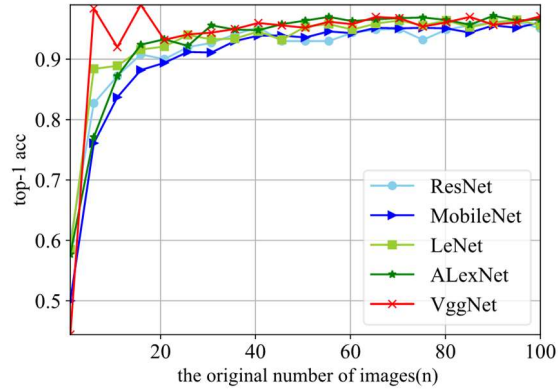

**Figure 5.** The relationship of $n$ and model accuracy on MNIST dataset ($n$ is the number of original images each class)

### 3.1.2. Omniglot and Analysis

The image classification results were evaluated by ND-MLS data augmentation on the Omniglot dataset that consists of 100 classes. Ten images were taken from each class for augmentation, and the rest of the data were used as a test set. Each image can generate 2004 new images by the ND-MLS method. The Omniglot dataset has 20 images for each class so that 2 million training images can be used for the training model. Let $C_n$ be the classes of original images in the Omniglot dataset. The final result can be obtained on 1,000 test images and evaluated by the top-1 error rates.

The performance of the models is evaluated when $C_n$=5. The results are shown in Table 3. All networks performed on the test set have a 0% top-1 error. This may be attributed to a limited number of test sets. To reveal the reasons, in Figure. 6, the factors of model accuracy should be observed by comparing the relationship between $C_n$ and model accuracy in the test set. First, the number of test images is directly proportional to $C_n$. When $C_n$ is less than 20, the test set is relatively small, making the accuracy fluctuate greatly. Second, when $C_n$ is greater than 20, the accuracy of VGGNet, ResNet, and AlexNet fluctuates slightly, and there is a small downward trend. It is concluded that the increase in classification has little effect on classification accuracy. Last, it is undeniable that models play a key role in the omniglot classification task. With the increase in $C_n$, the top-1 acc is notably affected by the structure of LeNet and MobileNet.

Furthermore, it is worth noting that the top-1 test error of VGGNet is 0.029, as Table 3 shows, and only ten images for each class were needed for ND-MLS data augmentation and model training. In other words, 97.1% top-1 acc for 100 different handwritten character classifications can be accomplished with only 1000 samples to be labeled. It is evident that small amounts of data can quickly be augmented by the ND-MLS method, and neural networks trained on them can obtain a good result.

**Table 3.** The top-1 error is on the test set of omniglot dataset and reported by the test server. $C_n$ is the classes of original images.

| | $C_n$=5 | $C_n$=10 | $C_n$=100 |
|---|---|---|---|
| LeNet | 0 | 0.05 | 0.151 |
| AlexNet | 0 | **0.03** | 0.048 |
| VggNet | 0 | **0.03** | 0.04 |
| ResNet | 0 | 0.04 | **0.035** |
| MobileNet | 0 | 0.06 | 0.111 |

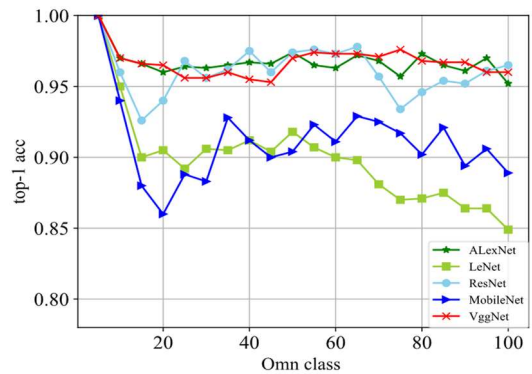

**Figure 6.** The relationship of $C_n$ and model accuracy on Omniglot dataset ($C_n$ is the classes of original images)

*3.2. ND-MLS performance in Semantic segmentation tasks*

Semantic segmentation tasks use a $\varphi_0$ of 45°, $\varphi_i = i*90°$ ($i=0,1,2,3$), and place the $p$ point on the edge of the object contour so that the deformation will be more realistic. Compared with the handwritten character problem in classification, semantic segmentation is much more sophisticated. In the handwriting case, the characters' positions in the image are mostly fixed. However, in cases where objects have complex contours, it is not practical to manually determine $p$ points because the label file does not contain the object contour information. Labeling is a time-consuming and labor-intensive task in semantic segmentation. Therefore, it will be important to augment the dataset using ND-MLS.

The object's shape and contours are deformed in many different ways by ND-MLS to augment the dataset. It is worth considering whether deformations lead to unrealistic training data. Figure 7 shows images of four different objects after deformation using different parameters. However, nine-dot MLS may suffer from fold backs, such as space warping approaches[16]. For example, Figure 7b (3), c (3), and d(2) are so distorted that they do not look real. [34] concludes that CNNs are strongly biased toward recognizing textures rather than shapes, which is in stark contrast to human behavioral evidence and reveals fundamentally different classification strategies. Therefore, it is determined that the distortion of object shape will not have a great impact on the training of the neural network.

It is worth exploring the textures of ND-MLS images. As shown in Figure 8, the original image is slightly different from the ND-MLS (local nonhomogeneity) deformed image, but in general, the deformed images display approximate spatial uniformity and consistency. When comparing the original and ND-MLS images, we can observe differences in partially enlarged details, the pixel arrangement, and their locations. ND-MLS changes the texture and position of the object, which can facilitate its perception by convolution networks.

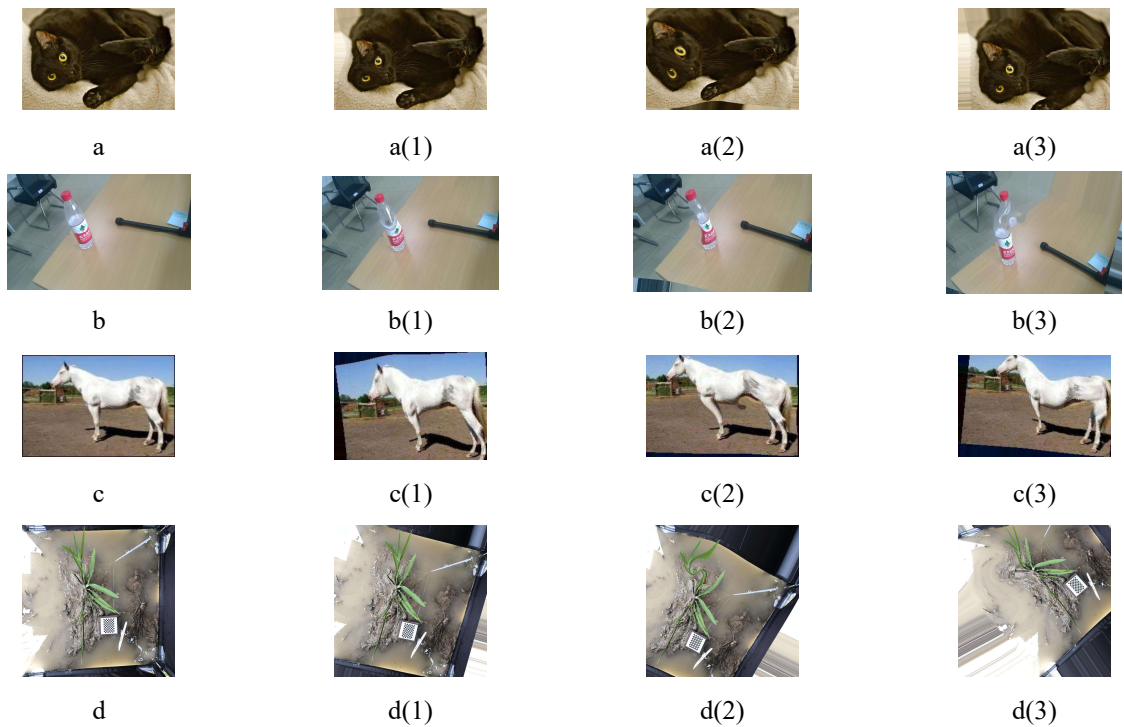

**Figure 7**. Original images (a, b, c, d) and its deformation using the ND-MLS method (a(1~3), b(1~3), c(1~3), d(1~3)).

### 3.2.1. Data collection and Analysis

The data set contains images of 4 object types. They can all be easily recognizable by eye. A dataset of 575 labeled images in total is formed. Among them, the Bombay cat is from The Oxford-IIIT Pet Dataset [35], and the horse is from a shared dataset. The bottle and grass dataset were created with the involvement of our team. A detailed list of this dataset is given in Table 4, and example images are given in Figure 7.

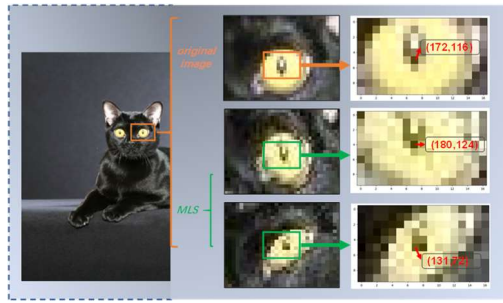

**Figure 8.** The partial enlarged detail of original &ND-MLS images. The coordinates are the positions of the pixels in the images.

The contrast between an object and its background can strongly influence the ability of the machine learning model to segment. Grass can show in different forms and colors, making it suitable for this ND augmentation method's verification. The Bombay cat and horse come from public data sets and are relatively more complex than the bottle and grass.

**Table 4.** The composition of the dataset.

|  | Original data set | ND-MLS train set | ND-MLS validation | Test set |
|---|---|---|---|---|
| bottle | 10+61 | 2004*8 | 2004*2 | 61 |
| grass | 10+52 | 2004*8 | 2004*2 | 52 |
| cat | 10+94 | 2004*8 | 2004*2 | 94 |
| horse | 10+328 | 2004*8 | 2004*2 | 328 |
| **Total** | 575 | 64128 | 16032 | 535 |

The ND-MLS dataset was evaluated on SegNet, PspNet, and DeepLab. They are all used for segmentations of one foreground object and the background. The original dataset contains 32 (train), 8 (validation), and 535 (test) pixel-level labeled images for training, validation, and testing, respectively. The dataset is augmented by ND-MLS resulting in 64128 training images and 16032 validated images. The performance is measured in terms of pixel intersection-overunion (IOU).

### 3.2.2. Results and Analysis

Table 5 shows the ND-MLS dataset's effect when employing SegNet, PspNet and DeepLab. As the number of original images increases, the m_Iou of the DeepLab model improves from 81.5% to 93.5%. The results show that (a) the DeepLab method is generally better than SegNet and PspNet. (b) The m_Iou based on 10 original images for DeepLab is 93.5%. It is amazing to obtain such a high score with only 10 original images. (c) The number of original images can improve the performance of bottle segmentation. It is obvious that the ND-MLS augmented dataset only needs a small amount of original data, and the trained model can achieve fairly good performance.

**Table 5.** Performance on bottle test set. (1) (5) (10) is the number of original images before ND-MLS deformation.

| methods | m_Iou(1) | m_Iou(5) | m_Iou(10) |
|---|---|---|---|
| SegNet | 0.569 | 0.807 | 0.896 |
| PspNet | 0.671 | 0.667 | 0.876 |
| DeepLab | 0.815 | 0.825 | 0.935 |

As shown in Table 6, we adopted the grass MLS dataset for semantic segmentation. The PspNet model attains 20.5% m_Iou(10), compared to 73.3% m_Iou(10) of DeepLab and 24.4% m_Iou(1) of PspNet. No improvement was observed when adopting PspNet on this grass MLS dataset. We considered that the model should have obtained a higher accuracy since the grass has only one color (green) and large texture similarity. However, even the highest m_Iou(10) does not meet our expectations. To explore the reason, we compare the qualitative segmentation results of the highest accuracy in different models, as shown in Figure 9. First, the grass has a large outer contour, while its leaves are small. The mis-segmented area gathers in the contour. Secondly, the ability of the model itself has an impact. The contour of the grass segmented by SegNet is zigzag, not smooth. The segmentation of grass fails directly on the PspNet model. If a higher m_Iou is desired, it can be achieved by increasing the number of original images or improving the model to make it more suitable for segmenting small objects such as grass. The purpose of this study is only to verify the feasibility of the MLS data set and not to improve the model. Therefore, it is necessary to use objects with more complex textures, such as cat and horse images, to verify the feasibility of MLS datasets.

**Table 6.** Performance on grass test set. (1,5,10) is the number of original images before ND-MLS deformation.

| methods | m_Iou(1) | m_Iou(5) | m_Iou(10) |
|---------|----------|----------|-----------|
| SegNet  | 0.416    | 0.560    | 0.601     |
| PspNet  | 0.244    | 0.201    | 0.205     |
| DeepLab | **0.620** | **0.693** | **0.733** |

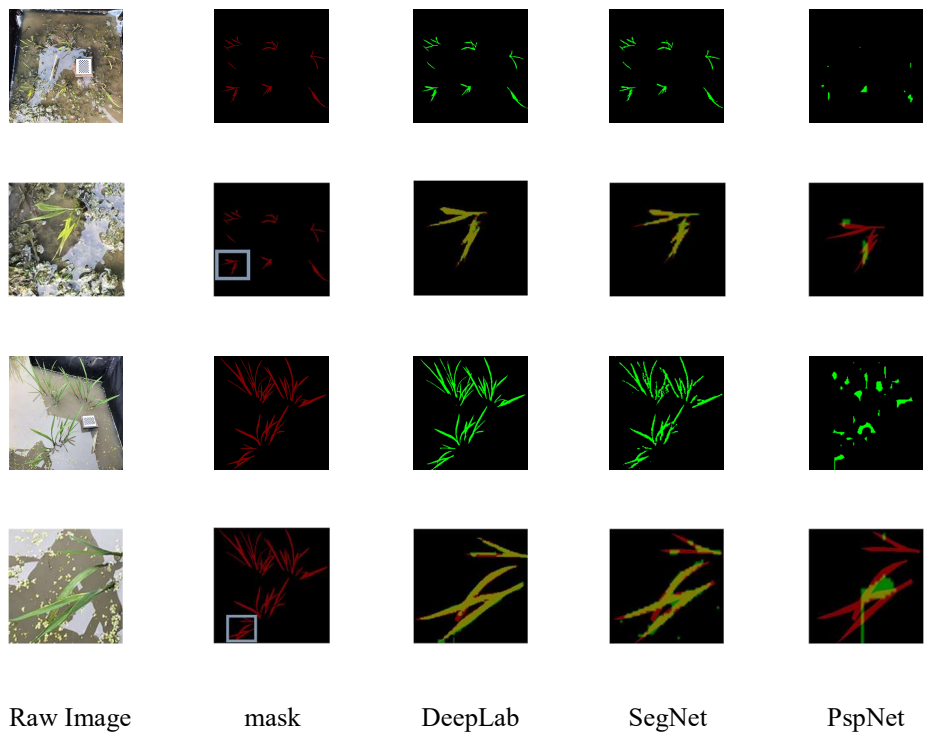

| Raw Image | mask | DeepLab | SegNet | PspNet |
|-----------|------|---------|--------|--------|

**Figure 9.** original image, ground-truth, and DeepLab SegNet PspNet results.

For the cat and horse, in m_Iou(1), the cat ND-MLS dataset alone yields 77.8%, significantly outperforming the horse by approximately 12.4%, as shown in Table 7 and 8. With black fur color, the Bombay cat is more accessible to segmentation than the horse when there is only one original image. Compared with the white horse-formed ND-MLS dataset, the black and brown horses have relatively difficulty gaining a higher m_Iou(1). In addition, the m_Iou(10) of the horse alone is 85%, significantly outperforming m_Iou(1) by 19.6%, but the cat results show only a 6.9% improvement. As the number of original samples increases, the horses' accuracy improves faster than that of Bombay cats. The horse has a similar segmentation effect as the cat by DeepLab when the number of original samples increases to 10. It can be concluded that the MSL dataset is relatively effective for training the segmentation model with only a small amount of data and achieves good segmentation results.

**Table 7.** Performance on bombay cat test set. (1,5,10) is the number of original images beform ND-MLS deform.

| methods | m_Iou(1) | m_Iou(5) | m_Iou(10) |
|---------|----------|----------|-----------|
| SegNet  | 0.729    | **0.830** | **0.867** |
| PspNet  | 0.732    | 0.794    | 0.835     |
| DeepLab | **0.778** | 0.803    | 0.847     |

**Table 8.** Performance on horse test set. (1) is the number of original images beform ND-MLS deform.

| methods | m_Iou(1) | m_Iou(5) | m_Iou(10) |
|---------|----------|----------|-----------|
| SegNet | 0.623 | 0.769 | 0.815 |
| PspNet | 0.419 | 0.618 | 0.763 |
| DeepLab | **0.654** | **0.785** | **0.850** |

To further analyze the ND-MLS dataset, we conducted experiments with more original images using different models. It can be seen from Figure 10a that the m_Iou curve of the segment model rises more smoothly as the number of original samples increases. The curve of PspNet in Figure. 10b has relatively high volatility and low accuracy, PspNet model has the worst performance, and even training failed on the grass data set, which may be attributable to the model's own problem. The curve of the DeepLab model is volatile in Figure 10c, and the initial accuracy is higher than that of the other models. Figure. 10d shows that each model's curves do not fluctuate significantly and tend to rise smoothly when the number of original samples is more than 6. Therefore, the DeepLab models should be used preferentially to obtain the highest accuracy when the number of original samples should be more than 6.

As shown in Figure 10(a~d), when there is only one original sample, the bottle and cat achieve better performance in DeepLab, while the accuracy of the others is low. After the number of original samples increases to 6, the segment accuracy of each type is high, and as the number continues to increase, the m_Iou growth rate slows significantly. When the number of original samples was 10, both SegNet and PspNet had more than 80% m_Iou in the cat and horse categories. DeepLab can reach over 90% on the bottle ND-MLS dataset. From the overall experimental data, only a small number of images are needed for ND-MLS augmentation, and semantic segmentation can achieve high accuracy.

**Current test result:** We uploaded the best model to the evaluation server, obtaining a performance of 70.4%. All models are tested on the original test set. The test set has not been augmented.

**Qualitative results:** We visualize the results in Figure 11.

### 3.2.3. ND-MLS performance in object detection tasks

The first stage is to find the contours of the mask in the ND-MLS dataset and then calculate the contour's outer rectangle. The coordinates of the outer rectangle form the object detection label. The Faster RCNN, YOLOv3, and YOLOv4 are used to verify object detection performance in the ND-MLS dataset. The performances of the cat, horse, and bottle datasets will be evaluated by m_Iou, AP50, and AP7.

Table 9 demonstrates the ND-MLS dataset's results on Faster RCNN, YOLOv3 and YOLOv4. The Faster RCNN model attains 63.6% AP75, compared to 18.1% AP75 of YOLOv3 and 40.9% AP50 of YOLOv4 when there is only one original image. The m_Iou and AP50 of Faster RCNN are also higher than those of YOLO v3 and YOLO v4. However, with the increase in the original ND-MLS images, the performance of YOLOv4 gradually exceeds that of Faster RCNN. When the original ND-MLS number is 10, YOLOv4 can attain 100% AP75. Under the premise of only a small number of samples, the YOLOv4 model has better accuracy on bottle detection by ND-MLS. Subsequently, we will evaluate the performance of cats and horses in object detection tasks.

As shown in Figure. 12a, the ND-MLS dataset is used to train the R_CNN model, and AP75 is positively correlated with the original number of images. In addition, the AP75 of cars and horses is relatively high even when the number of objects is only one, as shown in Figure. 12(b-c). As the number of original images increases, the AP75 of the horse and cat increases less but fluctuates more. Among the three models, the bottle performs best in YOLOv4, and the cat and horse models perform best in YOLOv4. As shown in Table 10, cats take 93.6% of AP75 (10) in YOLOv3, and horses take 94.8% of AP75 (5) in YOLOv3 and 97.2% AP75 (10) in YOLOv4. As shown in Table 10, the cat takes 93.6% of AP75 (10) in YOLOv3, the horse takes 94.8 % of AP75 (5) in YOLOv3, and 97.2 % AP75 (10) in YOLOv4 in Table 11. Thus, in single object detection, only a small amount of data needs to be labeled and augmented by ND-MLS to form a training set, and then the trained detection model can obtain good detection accuracy.

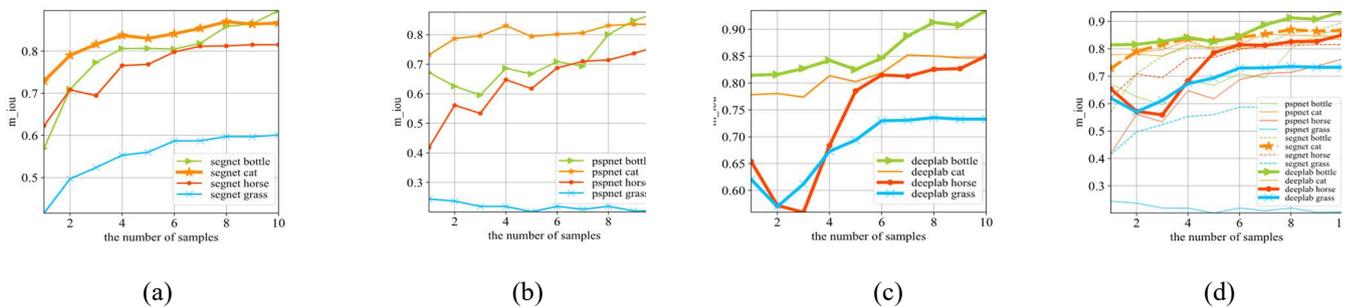

(a)       (b)       (c)       (d)

**Figure 10.** The relationship of sample's number and model accuracy on segmentation task (a) SegNet model; (b) PspNet; (c) DeepLab; (d) All model.

**Table 9.** Performance on the bottle test set. (1,5,10)is the number of original images before ND-MLS deformation.

| methods | m_Iou (1) | m_Iou(5) | m_Iou (10) | AP50 (1) | AP50 (5) | AP0 (10) | AP75 (1) | AP75 (5) | AP75 (10) |
|---|---|---|---|---|---|---|---|---|---|
| Faster RCNN | **0.707** | 0.787 | 0.790 | **0.732** | **0.954** | 0.984 | **0.636** | 0.863 | 0.909 |
| YOLOv3 | 0.262 | 0.725 | 0.827 | 0.409 | 0.909 | **1.000** | 0.181 | 0.773 | 0.864 |
| YOLOv4 | 0.521 | **0.829** | **0.859** | 0.590 | **0.954** | **1.000** | 0.409 | **0.909** | **1.000** |

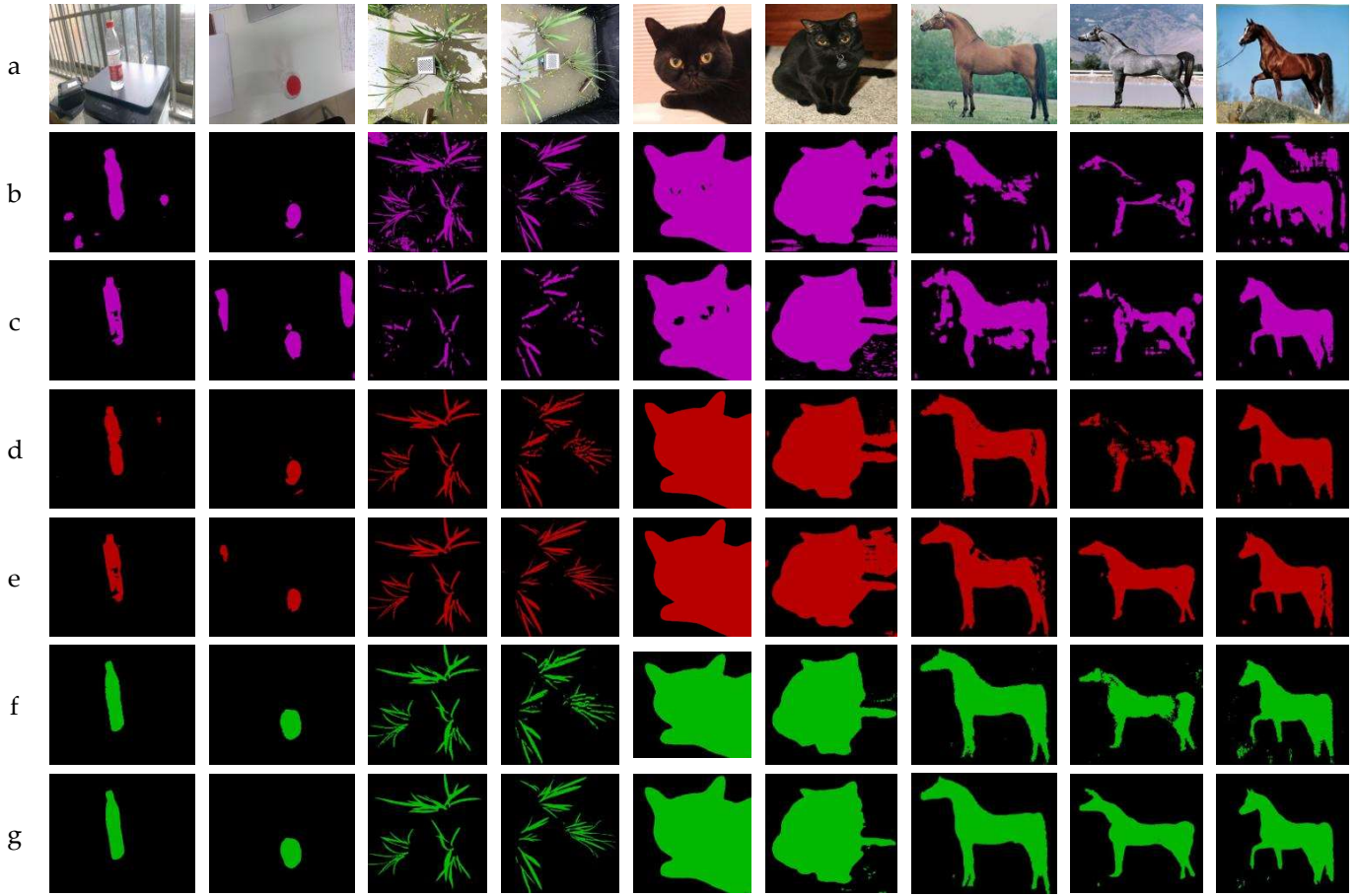

**Figure 11.** Visualization results on the ND-MLS data set when employing the DeepLab and SegNet model. the number of original images: 1 to pink, 5 to red,10 to green. (a) original images; (b, d, f) the mask in SegNet; (c, e, g) the mask in DeepLab

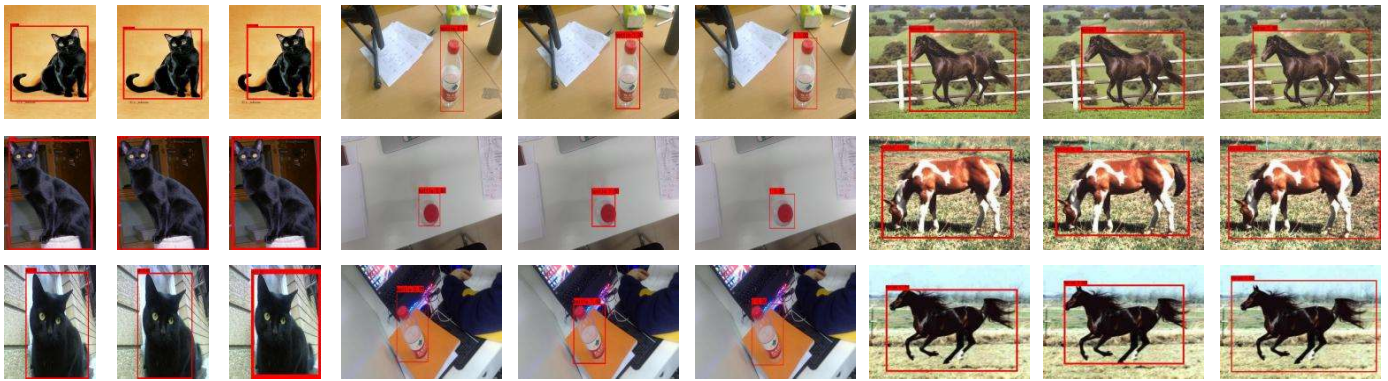

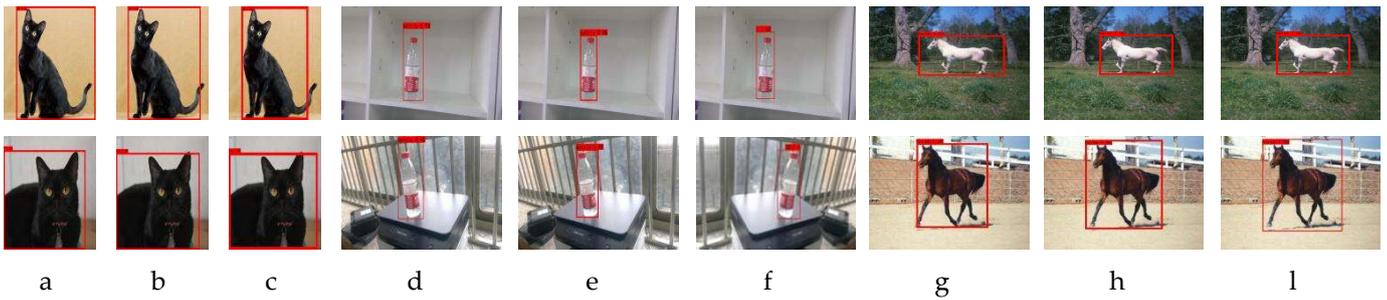

**Figure 12.** Visualization results on the ND-MLS data set when employing the YOLO v3, YOLO v4, and RCNN model by using ten original images (a,d,g) for YOLO v3, result (b, e, h) for YOLO v4, result (c,f,i) for RCNN result.

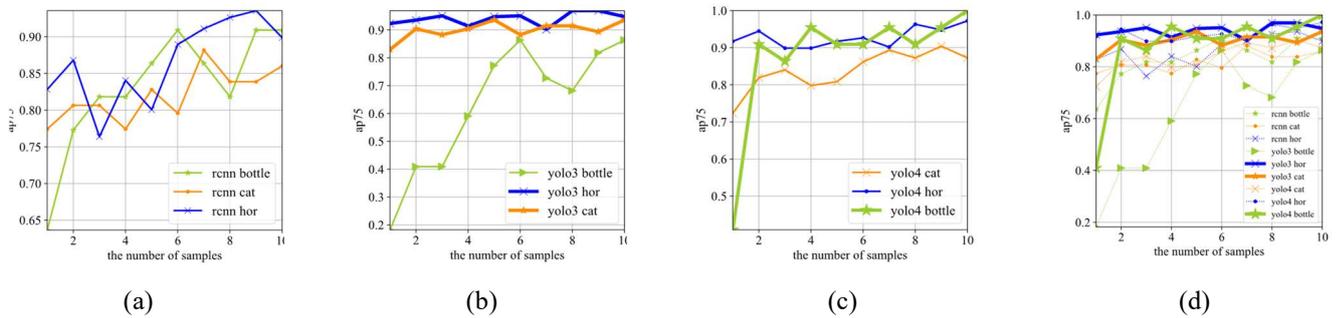

**Figure 13.** the relationship of sample's number and model accuracy on object detection task (a) RCNN; (b) YOLO v3; (c) YOLO v4; (d) All model.

**Current test result:** We tested the object detection model on the ND-MLS dataset, obtaining the best performance of 93.6% AP75(10) on the cat dataset, 97.2% AP75 (10) on the horse dataset, and 100% AP75(10) on the bottle dataset.

**Qualitative results:** We visualize the results in Figure 13, which shows object detection in the R-CNN, YOLOv3, and YOLOv4 models.

**Table 10.** Performance on cat test set.

| Methods | AP75(1) | AP75(5) | AP75(10) |
|---|---|---|---|
| Faster RCNN | 0.774 | 0.828 | 0.860 |
| YOLOv3 | **0.829** | **0.936** | **0.936** |
| YOLOv4 | 0.723 | 0.808 | 0.872 |

**Table 11.** Performance on horse test set.

| Methods | AP75(1) | AP7(5) | AP75(10) |
|---|---|---|---|
| Faster RCNN | 0.828 | 0.801 | 0.899 |
| YOLOv3 | **0.923** | **0.948** | 0.948 |
| YOLOv4 | 0.917 | 0.917 | **0.972** |

## 4. Conclusions

This paper introduces ND-MLS, a new approach for data augmentation. The feasibility of ND-MLS is tested in classification, object detection, and segmentation tasks. In terms of classification, LenNt, AlexNet, VGGNet, ResNet and MobileNet were used to verify the effect of ND-MLS in the MNIST dataset. Under the premise of only ten original pictures of each class of the dataset, the ND-MLS augmentation method can maintain the accuracy and attain 96.5 top-1 acc by ResNet on 100 different handwritten character classification tasks. VGGNet can gain 92% top-1 acc on the MNIST dataset of handwritten digits by ND-MLS.

In the segmentation problem, ND-MLS effects were verified using bottle, grass, cat, and horse datasets in the SegNet, PspNet, and DeepLab models. Only 10 original samples are used to train the model in each category after using ND_MLS, and the best results are as follows: DeepLab obtains 93.5%, 85%, and 73.3% m_IOU(10) in the bottle, horse, and grass test datasets, respectively, and the cat test dataset gains 86.7% m_IOU(10) in the SegNet model.

In the object detection problem, the ND-MLS effect of bottles, cats, and horses was verified in R-CNN, YOLO v3, and YOLO v4. With only 10 original samples in each category, the best AP75 values are as follows: YOLO v4 achieves 100% and 97.2% bottle and horse detection, respectively, and the cat dataset attains 93.6% on YOLO v3.

In summary, facing the challenge of limited data, few-shot augmentation (ND-MLS) methods can augment a large number of images using only a few labeled datasets quickly and obtain fairly good results.

This paper focuses on single object detection and segmentation verification for ND-MLS. The effect of multitarget detection and segmentation will be verified in the future. In this study, only a small number of labeled samples are needed to finish the detection and segmentation of objects. It is still difficult to train large models to cover many objects.

**Funding:** This research was funded by the National Science Foundation of China, grant number 61905219.

**Data Availability Statement:** In this section, The MNIST dataset could be found at http://yann.lecun.com/exdb/mnist/,The Omniglot dataset cound be found  at https:// gitee.com /CAVED/omniglot-dataset.